\documentclass{article}

\usepackage{microtype}
\usepackage{graphicx}
\usepackage{subcaption}
\usepackage{booktabs} %

\usepackage{hyperref}

\usepackage[ruled,vlined]{algorithm2e}

\newcommand{\attackfigsize}{0.13}
\newcommand{\trajfigsize}{0.245}

\usepackage[preprint]{neurips_2024}

\usepackage{amsmath}
\usepackage{amssymb}
\usepackage{mathtools}
\usepackage{amsthm}

\usepackage[capitalize,noabbrev]{cleveref}

\theoremstyle{plain}

\theoremstyle{definition}

\theoremstyle{remark}

\usepackage[textsize=tiny]{todonotes}

\usepackage[utf8]{inputenc} %
\usepackage[T1]{fontenc}    %
\usepackage{hyperref}       %
\usepackage{url}            %
\usepackage{booktabs}       %
\usepackage{multirow}       %
\usepackage{rotating}
\usepackage{makecell}
\usepackage{adjustbox}
\usepackage{amsfonts}       %
\usepackage{nicefrac}       %
\usepackage{microtype}      %
\usepackage{xcolor}         %
\usepackage{amsmath}
\usepackage{enumitem}
\usepackage{wrapfig}
\usepackage{adjustbox}
\usepackage[normalem]{ulem}

\setcitestyle{numbers,square,sorted}

\title{Protecting against simultaneous data poisoning attacks}

\begin{document}

\author{Neel Alex\thanks{Correspondence to \texttt{sma92@cam.ac.uk}} \\
        University of Cambridge \\
        \And
        Shoaib Ahmed Siddiqui \\
        University of Cambridge \\
        \AND
        Amartya Sanyal \\
        Max Planck Institute, Tübingen \\
        \And
        David Krueger \\
        University of Cambridge \\
}

\maketitle

\begin{abstract}
    Current backdoor defense methods are evaluated against a single attack at a time. This is unrealistic, as powerful machine learning systems are trained on large datasets scraped from the internet, which may be attacked multiple times by one or more attackers. We demonstrate that simultaneously executed data poisoning attacks can effectively install multiple backdoors in a single model without substantially degrading clean accuracy. Furthermore, we show that existing backdoor defense methods do not effectively prevent attacks in this setting. Finally, we leverage insights into the nature of backdoor attacks to develop a new defense, BaDLoss, that is effective in the multi-attack setting. With minimal clean accuracy degradation, BaDLoss attains an average attack success rate in the multi-attack setting of 7.98\% in CIFAR-10 and 10.29\% in GTSRB, compared to the average of other defenses at 64.48\% and 84.28\% respectively.\footnote{We open-source our code to aid replication and further study, available on GitHub: \url{https://github.com/shoaibahmed/mapd_backdoors.}}
\end{abstract}

\section{Introduction}
Many deep learning applications use large-scale datasets obtained through web scraping with minimal curation. These datasets are vulnerable to attackers, who can easily inject data that alters the behavior of models trained on these datasets. ~\citet{carlini2023poisoning} demonstrated that poisoning real-world, large-scale datasets is a feasible threat due to their distributed nature.

Among the various data poisoning threats, the creation of model backdoors is particularly insidious. By modifying only a small number of examples in a dataset, adversaries can make a trained model sensitive to highly specific features. The adversary can then control the model's outputs by injecting these features into otherwise innocuous images~\citep{gu2017badnets,liu2018trojan,barni2019sinusoid,li2021antibackdoor,carlini2023poisoning} -- despite the model appearing benign during regular evaluation.

To the authors' knowledge, all prior works only evaluate against one backdoor attack at a time ~\citep{gu2017badnets,liu2018trojan,barni2019sinusoid,li2021antibackdoor}.
Real-world poisoned data sets can potentially include multiple backdoors of different types, as large-scale distributed datasets~\cite{carlini2023poisoning} have a very low barrier for adding data, so multiple attackers could easily add poisoned examples aimed at installing multiple backdoors. Even a single attacker could deploy multiple attacks. This setting is both more realistic and makes it substantially more challenging to defend against attacks in practice.

Therefore, in this work we propose the problem setting of defending against \textbf{\emph{multiple simultaneous backdoor attacks}}. We demonstrate that a poisoned dataset can introduce multiple backdoors into a target model without substantial clean accuracy degradation, and that existing defenses fail to effectively defend in this setting. As an additional contribution, we identify a common property of poisoned images in datasets -- that they demonstrate anomalous loss trajectories as a result of using unnatural features for classifcation. Correspondingly, we propose a defense called \textbf{BaDLoss} meant to be robust to multiple simultaneous backdoor attacks. We demonstrate that new defense performs well compared to other defenses, and retains performance in the synthetic single-attack setting.

\section{Related Work}

Since our work is concerned with backdoor detection and prevention, we will briefly discuss both attacks as well as defenses presented in the past.
Most work on backdoors, including ours, focuses on image datasets, although several recent works also explore poisoning of language models \citep{wan2023poisoning, hubinger2024sleeper}.

\subsection{Backdoor Attacks}

While attacks exist where the adversary provides a secretly backdoored model to the victim \citep{nguyen2020inputaware, tan2020adversarialembeddings}, we focus on the \textbf{data poisoning} setting, where the attacker manipulates a subset of the data to later control the model by injecting a feature of their choice.

Early data poisoning attacks modified both training images and their corresponding labels, adding a small stamp~\citep{gu2017badnets}, or overlaying an image~\citep{chen2017backdoor} or pattern~\citep{liao2018invisible}, then changing the label of altered images to a \textbf{target class}, so the model associates the injected feature with the target class. Later attacks invisibly warp images~\citep{nguyen2021wanet} or overlay image-specific invisible perturbations~\citep{li2021invisible}. Some attacks do not even change images' labels. Clean-label attacks alter training images with features that the learning algorithm preferentially detects~\citep{turner2019labelconsistent,barni2019sinusoid} without changing any image classes. More sophisticated strategies use realistic reflection overlays~\citep{liu2020reflection} or random-noise patterns~\citep{souri2022sleeper, zeng2022narcissus} that subtly induce the model to learn the desired behavior.

\subsection{Backdoor Defenses}

Various defense mechanisms have been proposed to mitigate model backdoors. One approach, known as trigger reverse engineering, assumes that a trigger mask with very few pixels causes the model to misbehave. Neural Cleanse~\citep{wand2019neuralcleanse} and variants ~\citep{guo2019tabor,tao2022pixelbackdoor,dong2021blackbox,wang2020trojannet} learn the trigger mask from a trained model, while others employ techniques like activation patching~\citep{liu2019abs} or generative modeling~\citep{qiao2019triggerdistribution} to identify the trigger, after which neurons which activate on the trigger can be pruned, or a number of other cleaning methods can be applied.

Backdoors can also sometimes be detected by directly examining model activations. Activation Clustering and Spectral Signatures~\citep{tran2018spectral} assume clean and backdoored examples demonstrate distinctive patterns in the model's activation space. Further work examines specific properties of activation space and identifies neurons that highly influence model behavior on backdoored examples ~\citep{zheng2022preactivation,wu2021adversarialpruning,xu2021linfinitypruning}. These neurons can then be pruned~\citep{liu2018finepruning}, or the model fine-tuned appropriately~\citep{chen2019deepinspect,li2021nad,zeng2022ibau}.

As removing backdoors from a trained model is challenging~\citep{goel2022adversarial}, many defenses indirectly prevent the backdoor from being learned at all~\citep{hong2020dpsgdpoisoning,borgnia2021dpaugmentations,huang2022decoupling, wang2022trapandreplace, chen2022poisonedsensitivity}. Anti-Backdoor Learning~\citep{li2021antibackdoor} \textit{maximizes} the loss of identified poisoned examples during training, making the model ignore poisoned features, a method used by other detection methods \citep{huang2023cognitivedistillation}. Finally, other approaches detect inputs that cause misbehavior in the output of the models~\citep{gao2020strip,chou2020sentinet,kiourti2021misa} or exploit the unique frequency spectra of attack to identify attack inputs~\citep{zeng2022frequency}, then filter them out for retraining or at deployment.

\section{Multiple Simultaneous Attacks}

To the authors' knowledge, no prior work evaluates on multiple backdoor attacks simultaneously. Therefore, we will begin with a demonstration that this setting is feasible in practice without substantial clean-accuracy degradation. 

We compare against a wide variety of classic and state-of-the-art backdoor attacks and defenses in the literature. Following recent work \citep{wang2022trapandreplace, kiourti2021misa, li2021nad, do2023inputfiltering}, we use the standard computer vision datasets: CIFAR-10~\citep{krizhevsky2009cifar} and GTSRB~\citep{houben2013gtsrb}, which capture variance both in image size (32x32 for CIFAR-10, versus variable image sizes up to 250x250 for GTSRB) as well as class distribution (CIFAR-10 has 10 balanced classes, GTSRB has 43 imbalanced classes). 
We use CIFAR-10 as a development set to tune defenses and, importantly, apply the tuned methods directly to GTSRB \textit{without any further tuning} for a more accurate assessment of the methods' performance in realistic settings.

\subsection{Threat Model}

In our threat model, we assume that the attacker(s) have control only over a portion of the training dataset which they can view and modify. The size of this modified fraction is referred to as the poisoning ratio $p$. The defender has complete control over the training process, and the attacker has no control. However, the defender receives a labeled dataset from an unvalidated external source. This corresponds well to an attack setting in which the victim trains a model using their own internal, well-tested code, but relies on a large, externally sourced dataset that cannot be manually quality-checked. These assumptions align with threat models commonly considered in prior works~\citep{carlini2023poisoning,chen2018activationclustering,li2021antibackdoor,do2023inputfiltering}.

Additionally, we assume the defender has access to a small set of guaranteed clean examples~(here 250 examples). Many defense methods~\citep{liu2018finepruning, wand2019neuralcleanse, gao2020strip, chou2020sentinet, kiourti2021misa,  li2021nad, zeng2022frequency, wang2022trapandreplace, huang2022decoupling} assume access to bona fide clean examples, as trusted human labor could manually generate or filter a subset of the dataset.

When evaluating, we primarily evaluate retrained model's accuracy on the clean test set, and the success rate of any attacks present in the data on the retrained model. This is consistent with defense evaluations considered in prior works~\cite{li2021antibackdoor,chen2018activationclustering,wand2019neuralcleanse}.We define clean accuracy as the retrained model's performance on the test set. Attack success rate (ASR) is evaluated on the full test set \textit{excluding} the target class, with the backdoor injected into every example to verify that the attack successfully changes the model's predictions. Full training details are available in Section~\ref{sec:train-details} in the Appendix.

\subsection{Attacks Considered}

\begin{figure}[!t]\centering
    \begin{subfigure}[b]{\attackfigsize\textwidth}
         \centering
        \includegraphics[width=\textwidth]{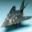}
        \caption{}
    \end{subfigure}
    \begin{subfigure}[b]{\attackfigsize\textwidth}
         \centering
        \includegraphics[width=\textwidth]{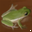}
        \caption{}
    \end{subfigure}
    \begin{subfigure}[b]{\attackfigsize\textwidth}
         \centering
        \includegraphics[width=\textwidth]{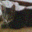}
        \caption{}
    \end{subfigure}
    \begin{subfigure}[b]{\attackfigsize\textwidth}
         \centering
        \includegraphics[width=\textwidth]{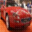}
        \caption{}
    \end{subfigure}
    \begin{subfigure}[b]{\attackfigsize\textwidth}
         \centering
        \includegraphics[width=\textwidth]{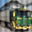}
        \caption{}
    \end{subfigure}
    \begin{subfigure}[b]{\attackfigsize\textwidth}
         \centering
        \includegraphics[width=\textwidth]{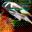}
        \caption{}
    \end{subfigure}
    \begin{subfigure}[b]{\attackfigsize\textwidth}
         \centering
        \includegraphics[width=\textwidth]{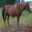}
        \caption{}
    \end{subfigure}
    \vspace{-3mm}
    \caption{\small Complete list of attacks considered in this work. \textbf{(a)} checkerboard pattern trigger (Patch)~\cite{gu2017badnets}, \textbf{(b)} single pixel trigger (Single-Pix)~\cite{gu2017badnets}, \textbf{(c)} random noise blending attack (Blend-R)~\cite{chen2017backdoor}, \textbf{(d)} dimple pattern blending attack (Blend-P)~\citep{liao2018invisible}, \textbf{(e)} sinusoid pattern blending attack (Sinusoid)~\citep{barni2019sinusoid}, 
    \textbf{(f)} optimized-trigger attack (Narcissus)~\citep{zeng2022narcissus}, and \textbf{(g)} frequency-domain attack (Frequency)~\citep{wang2021frequencybackdoor}}
    \label{fig:attacks}
\end{figure}

All considered attacks are visualized in Figure~\ref{fig:attacks}. All attacks change the images' labels except for the sinusoid attack, which is a clean-label attack. All attacks target different labels, and no image is attacked twice (so each image contains at most one trigger). The Narcissus \citep{zeng2022narcissus} attack is not used in GTSRB, as the optimized trigger is challenging to generate on variable-sized images.

In this work, we evaluate against the following attacks:
\begin{itemize}
    \item \textbf{Patch}: Adds a small patch in the corner of attacked images \citep{gu2017badnets}.
    \item \textbf{Single-Pixel}: Adds a single pixel in the corner of attacked images \citep{gu2017badnets}.
    \item \textbf{Blend-Random}: Blends a random noise pattern with attacked images \citep{chen2017backdoor}.
    \item \textbf{Blend-Pattern}: Blends a dimple pattern with attacked images \citep{liao2018invisible}.
    \item \textbf{Frequency}: Adds peaks in attacked image's discrete cosine transform \citep{wang2021frequencybackdoor}.
    \item \textbf{Sinusoid}: Adds sinusoidal stripes to attacked images of a single class \citep{barni2019sinusoid}.
    \item \textbf{Narcissus}: Adds a learned trigger to a very small number of attacked images \citep{zeng2022narcissus}.
\end{itemize}
In the multi-attack setting, all attacks are present at their full poisoning ratio. Full attack specifications are available in Section~\ref{sec:attacks} in the Appendix .

\subsection{Efficacy of Multiple Attacks}

\begin{figure*}[t]
    \centering
        \includegraphics[width=0.9\textwidth]{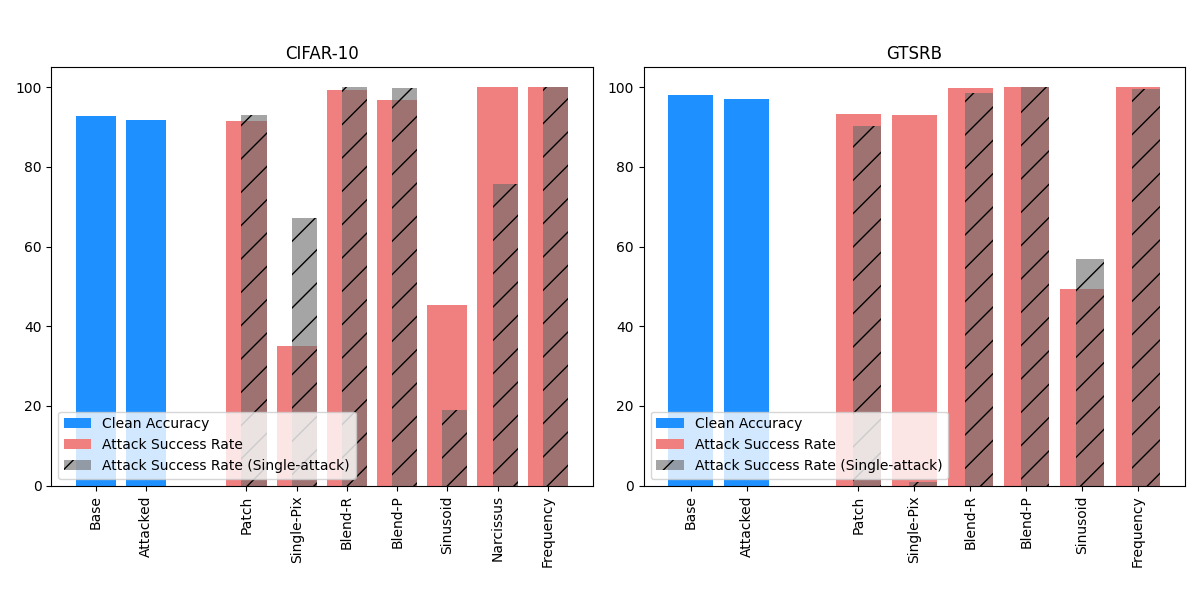}
    \caption{\small \textbf{Evaluation of Multi-attack Setting.} Clean accuracy (blue) should be high, attack success rate (red) should be low. When multiple attacks are simultaneously deployed, clean accuracy suffers slightly but is generally preserved at a high level. All attacks are simultaneously learned by the final model. 
    We also plot the attack success rate against single attacks to illustrate positive and negative interference between attacks.}
    \label{fig:multiattack}
\end{figure*}

Our results are visualized in Figure~\ref{fig:multiattack} (exact numerical results available in Table~\ref{table:multi-attack} in the Appendix). In both CIFAR-10 and GTSRB, the presence of multiple attacks does not substantially degrade clean accuracy. Additionally, in both settings, all attacks are learned simultaneously to a reasonably high degree. Notably, some attacks changed in performance substantially between the single-attack and multi-attack setting, such as the single-pixel attack, which is completely unlearned when isolated in GTSRB, but which is learned to over 90\% attack success rate in the multi-attack setting. This indicates that even attack-development research should consider the multi-attack setting, as it can have dramatic impacts on an attack's viability in realistic threat models.

We take these results as validating the importance of the multi-attack setting. If one or more attackers, by controlling small parts of a large, internet-scraped dataset, can cause a target model to respond to many different types of backdoor triggers without substantially degrading performance on clean data in a way that a defender might notice, then this setting is an vitally important one. However, existing defenses may be able to effectively adapt to this new setting.

\subsection{Defenses Considered}

In this work, we evaluate the following defenses:
\begin{itemize}[leftmargin=0.2in]
    \item \textbf{Neural Cleanse} \citep{wand2019neuralcleanse} Neural Cleanse learns a minimum-magnitude mask per class to reclassify every image in the dataset as that class. Masks with substantially-below-median magnitude are considered backdoor triggers.
    \item \textbf{Activation Clustering} \citep{chen2018activationclustering} Activation Clustering decides that a backdoor is present if the last layer's activations for a given class can be clustered into two classes, then removes the smaller cluster for retraining.
    \item \textbf{Spectral Signatures} \citep{tran2018spectral} Spectral signatures uses a singular value decomposition of last layer activations per class, and removes the 15\% of data with the highest value in the first singular dimension for retraining.
    \item \textbf{Frequency Analysis} \citep{zeng2022frequency} Frequency analysis identifies poisoned examples by building a classifier on the discrete cosine transforms of synthetic images with a fixed set of hardcoded backdoor-like features.
    \item \textbf{Anti-Backdoor Learning} \citep{li2021antibackdoor} Anti-backdoor learning identifies examples with particularly low loss after a few epochs of training and removes them before continuing training. After training is concluded, a few epochs of loss maximization on those examples removes sensitivity to the backdoor.
    \item \textbf{Cognitive Distillation} \citep{huang2023cognitivedistillation} Cognitive distillation learns a mask for each example that yields the same activations, treating examples with particularly low magnitude as backdoors, then uses ABL's unlearning method.
\end{itemize}
Full defense specifications are available in Section~\ref{sec:defenses} in the Appendix.

\subsection{Existing Defense Performance}

\begin{figure*}[t]
    \centering
        \includegraphics[width=0.9\textwidth]{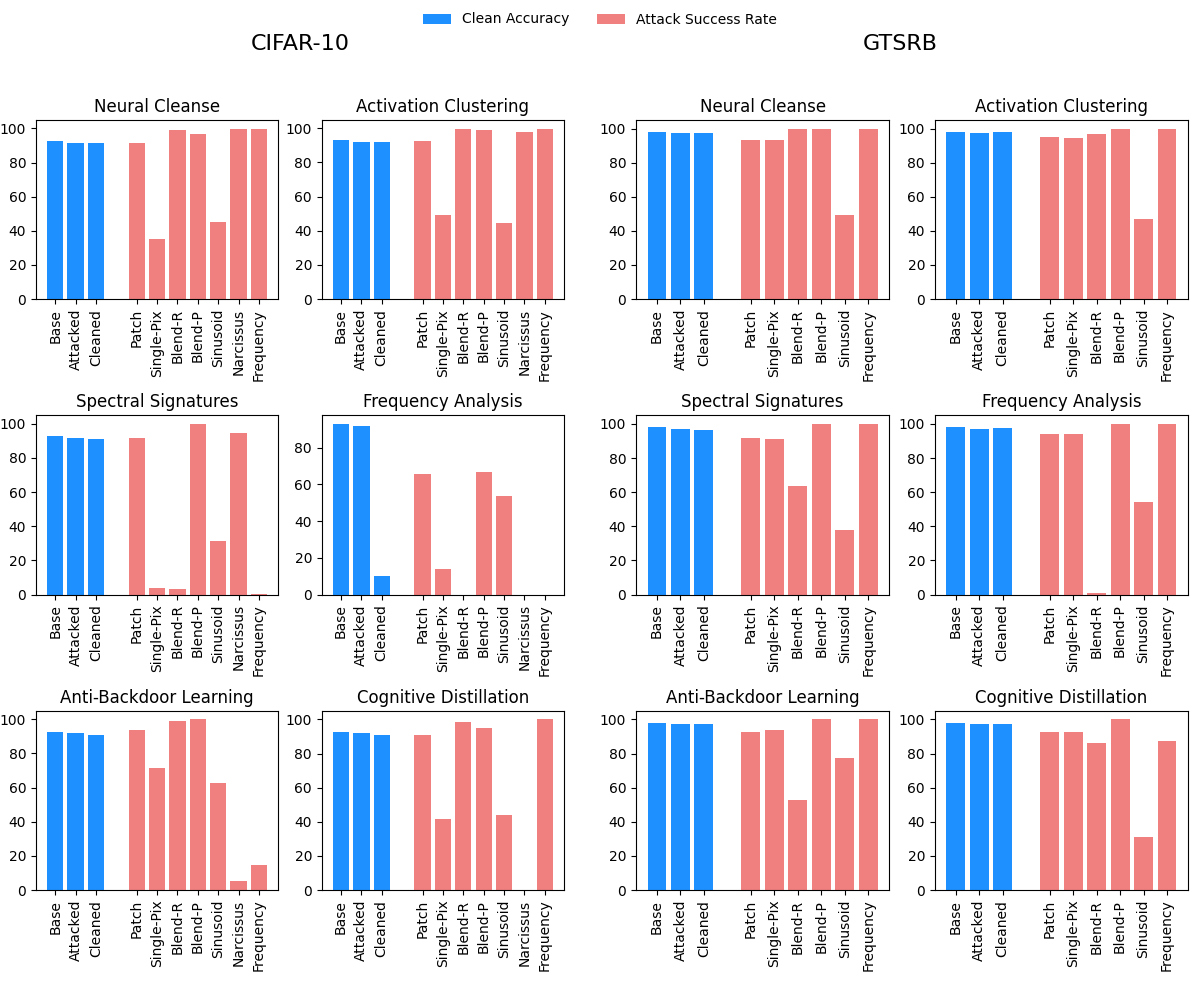}
    \caption{\textbf{Defense Performance in the Multi-attack Setting.} All evaluated defenses demonstrate failures when evaluated in the multi-attack setting.  This holds across datasets.
        The frequency analysis defense method \citep{zeng2022frequency}  achieves low attack success rate on CIFAR-10, but removes so much data that clean accuracy is reduced to random chance levels.
        Spectral Signatures provides the best defense overall on CIFAR-10, but fails on GTSRB, and dramatically underperforms BaDLoss (see Figure~\ref{fig:multiattack-badloss})}
    \label{fig:multiattack-defense}
\end{figure*}

Our results are visualized in Figure ~\ref{fig:multiattack-defense} (exact numerical results available in Table~\ref{table:multi-attack} in the Appendix). Except for the frequency analysis defense in CIFAR-10, which discards so much data that the clean accuracy is irrepairably degraded, all defenses in the multi-attack setting fail to defend against a variety of attacks. Even the \textit{best} performing defense, Spectral Signatures in CIFAR-10, has three attacks performing at nearly-perfect attack success rate.

Likely due to making assumptions incompatible with the complexity of the multi-attack setting, current backdoor defenses, when used with their default parameters, do not effectively defend against these relatively standard attacks when deployed simultaneously. In order to make progress in the multi-attack setting, new defenses must be developed.

\section{Developing a New Defense}

\begin{figure*}[t]
    \centering
        \includegraphics[width=0.9\textwidth]{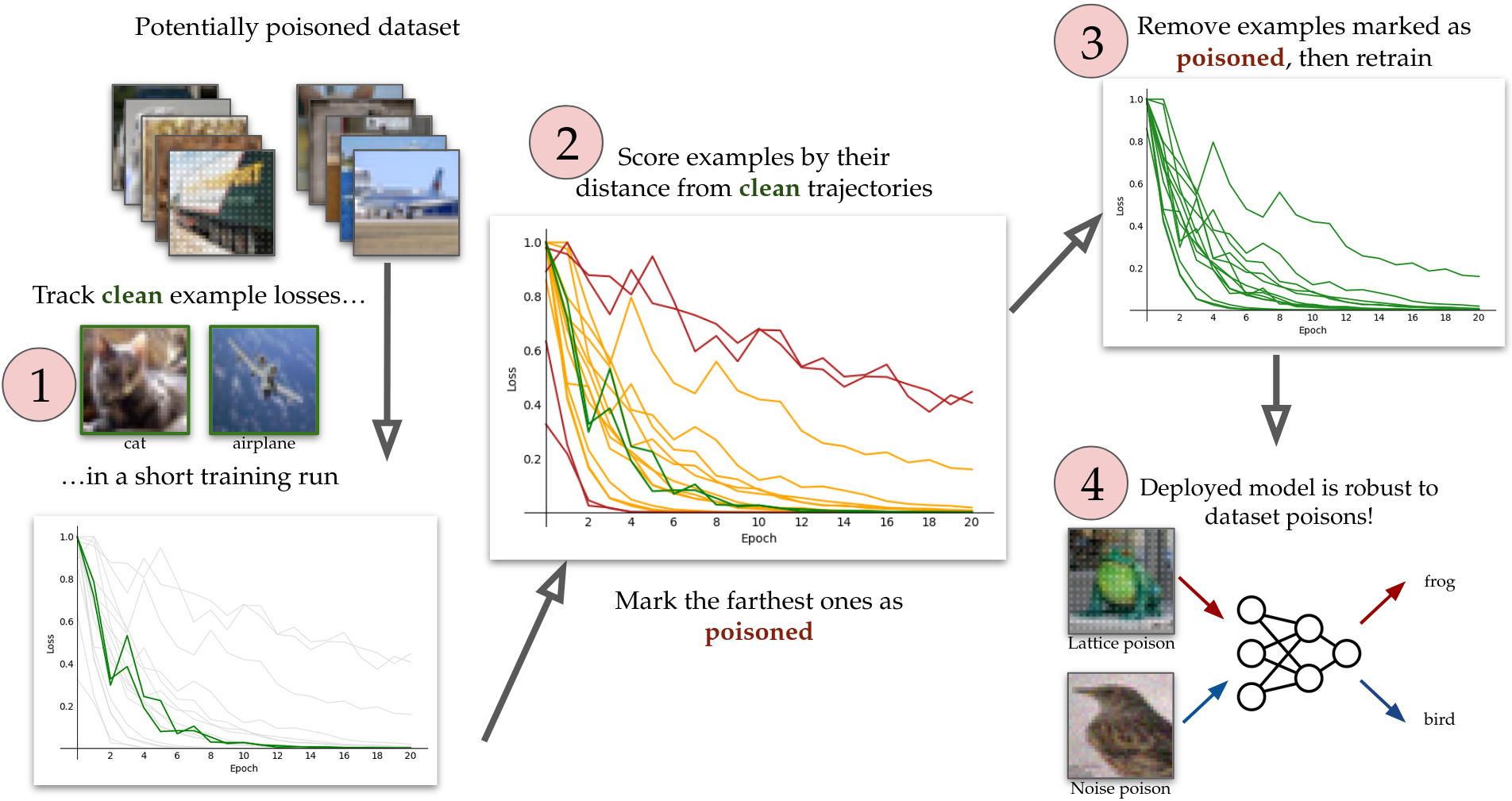}
    \caption{\small \textbf{BaDLoss Overview.} 
    (1) The defender tracks clean examples (``probes'') in the training set across multiple short training runs. (2) Every example gets an anomaly score based on its average distance from the bona fide clean examples. The farthest examples are marked as potential backdoors. (3) The defender retrains the model, excluding any examples identified as anomalous. (4) The defender deploys the more robust model.}\vspace{-5pt}
    \label{fig:badloss_overview}
\end{figure*}

To defend in the multi-attack setting, we first note that many assumptions made by existing defenses, while accurate in the single-attack setting, no longer hold when multiple simultaneous attacks occur. For instance, neural cleanse assumes that the attacked class will have an anomalously low mask magnitude relative to the other classes, but if sufficiently many classes are attacked, the median mask magnitude may be low enough that no classes look anomalous.

We look to attack-agnostic metrics that could indicate the presence of a backdoor. Several methods exist that analyze the loss of training examples to determine whether such examples are installing a backdoor. Anti-Backdoor Learning assumes that backdoored examples achieve a lower loss more quickly than normal examples~\citep{li2021antibackdoor}, and \citet{khaddaj2023rethinking} similarly claim that backdoor images have the \textit{strongest} features and are thus learned much more easily. Other works~\citep{hayase2021spectre} treat backdoors as anomalous examples which conflict with natural image features, and are therefore harder to learn. Such claims are correct in context, but miss the whole picture. We synthesize these ideas: images can exhibit either faster or slower training dynamics compared to a dataset's natural images, depending on the nature of the backdoor. As a defender cannot know what backdoor methods will be deployed, a defense must be effective against backdoors that induce \textit{any} unusual training dynamics.

Therefore, we introduce \textit{\textbf{BaDLoss} (Backdoor Detection via Loss Dynamics)}, visually demonstrated in Fig.~\ref{fig:badloss_overview}. To generalize beyond assumptions that all backdoors are simply easier or harder to learn, BaDLoss leverages a small number of bona fide clean examples that are used as part of the training set. A model is subsequently trained using all available examples, including the reference clean examples, while simultaneously tracking the loss associated with each instance in the training set. The added bona fide clean examples provide reference trajectories that allow us to define an anomaly score for each example in the training set based on the distance to the clean trajectories.

\vspace{-5pt}
\subsection{BaDLoss: Backdoor Detection via Loss Dynamics}

Our approach draws inspiration from the Metadata Archaeology via Probe Dynamics (MAP-D) technique~\citep{siddiqui2022metadata}, which examines per-example loss trajectories in a training set to remove unusual or mislabelled examples. Notably, other studies have also harnessed training dynamics for various objectives~\citep{kaplun2022deconstructing,liu2022membership,rabanser2022selective}.

\begin{figure*}[t]\centering
    \begin{subfigure}[b]{\trajfigsize\textwidth}
         \centering
        \includegraphics[width=\textwidth]{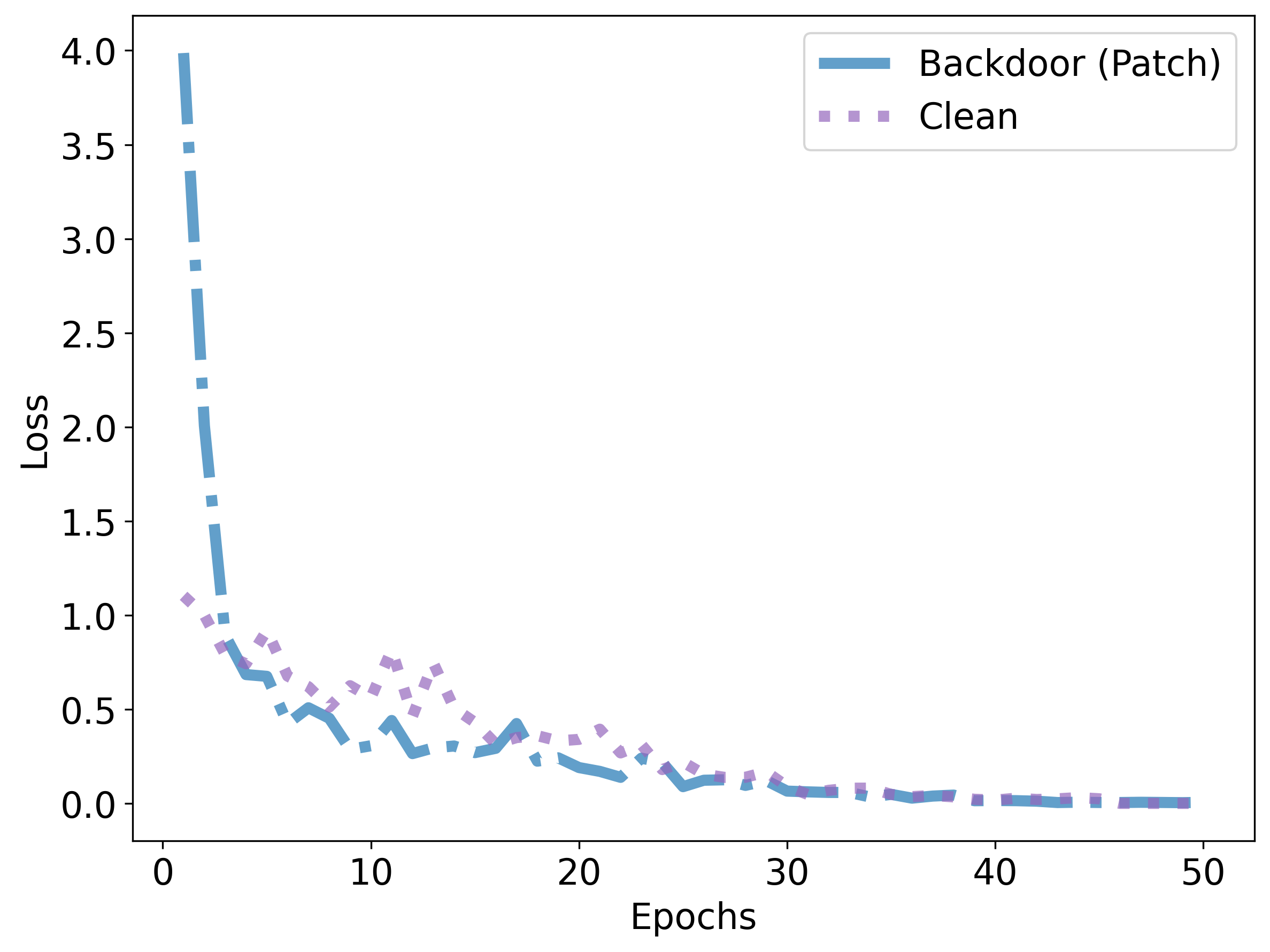}
    \end{subfigure}
    \begin{subfigure}[b]{\trajfigsize\textwidth}
         \centering
        \includegraphics[width=\textwidth]{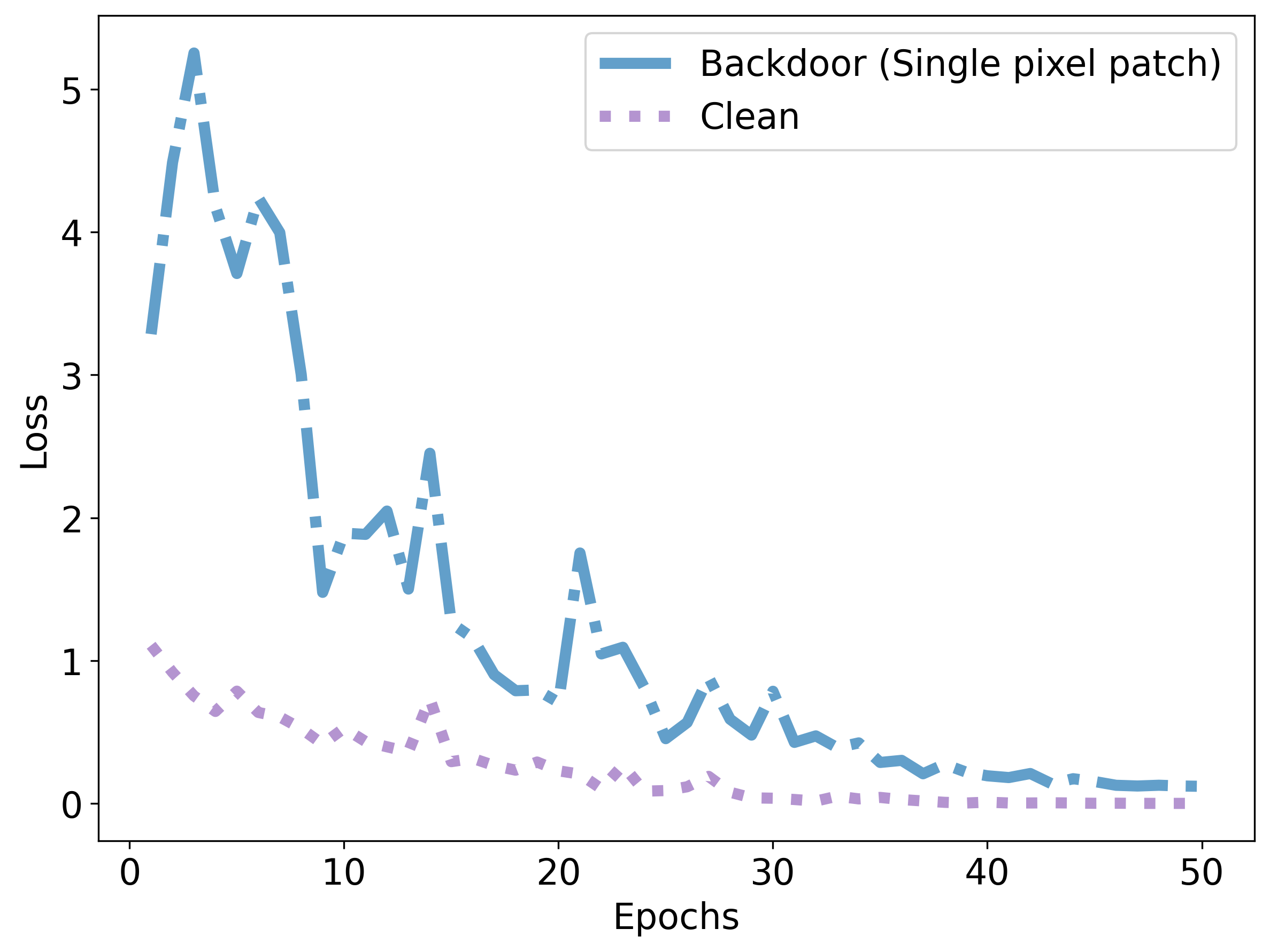}
    \end{subfigure}
    \begin{subfigure}[b]{\trajfigsize\textwidth}
         \centering
        \includegraphics[width=\textwidth]{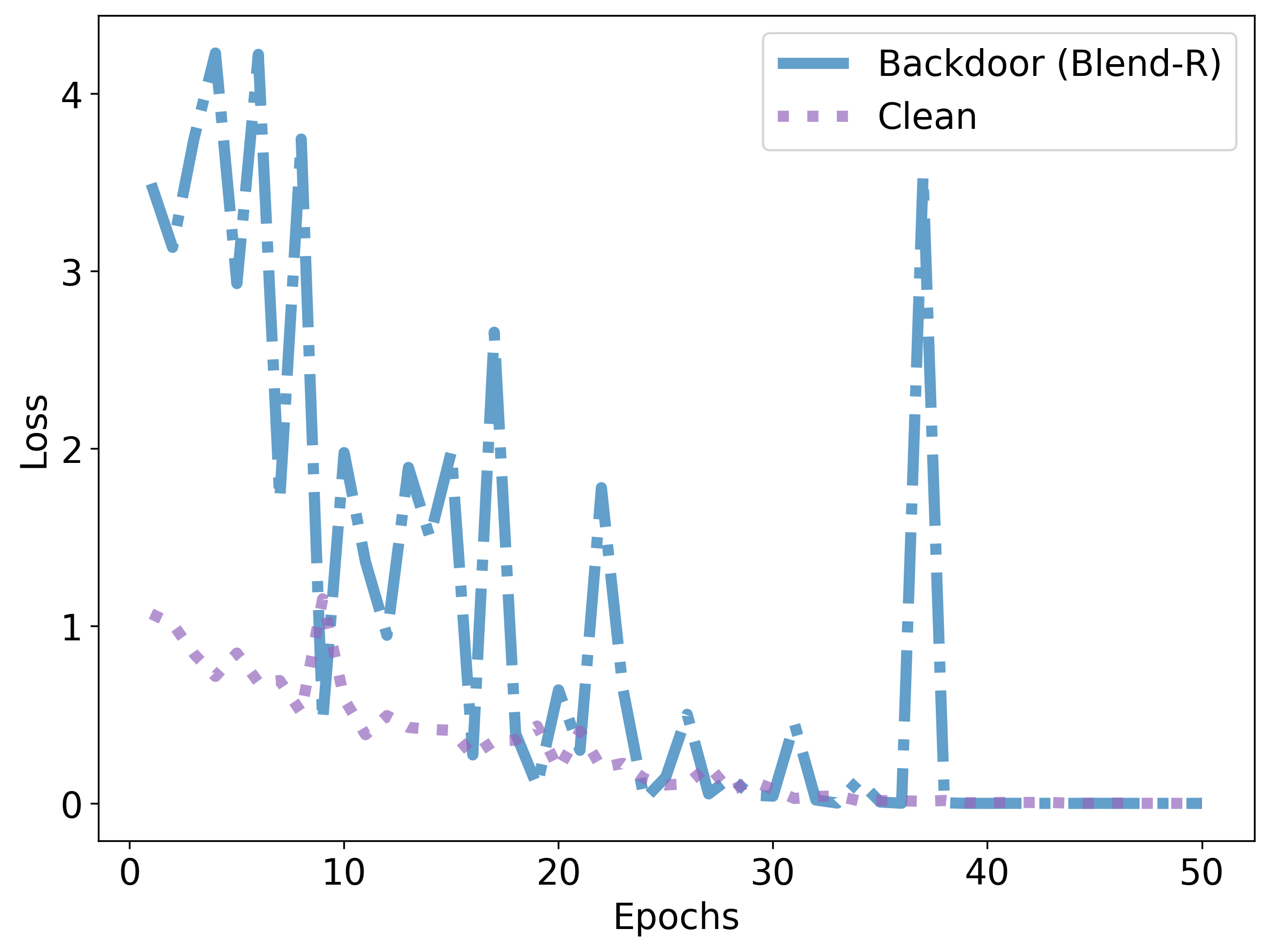}
    \end{subfigure}
    \begin{subfigure}[b]{\trajfigsize\textwidth}
         \centering
        \includegraphics[width=\textwidth]{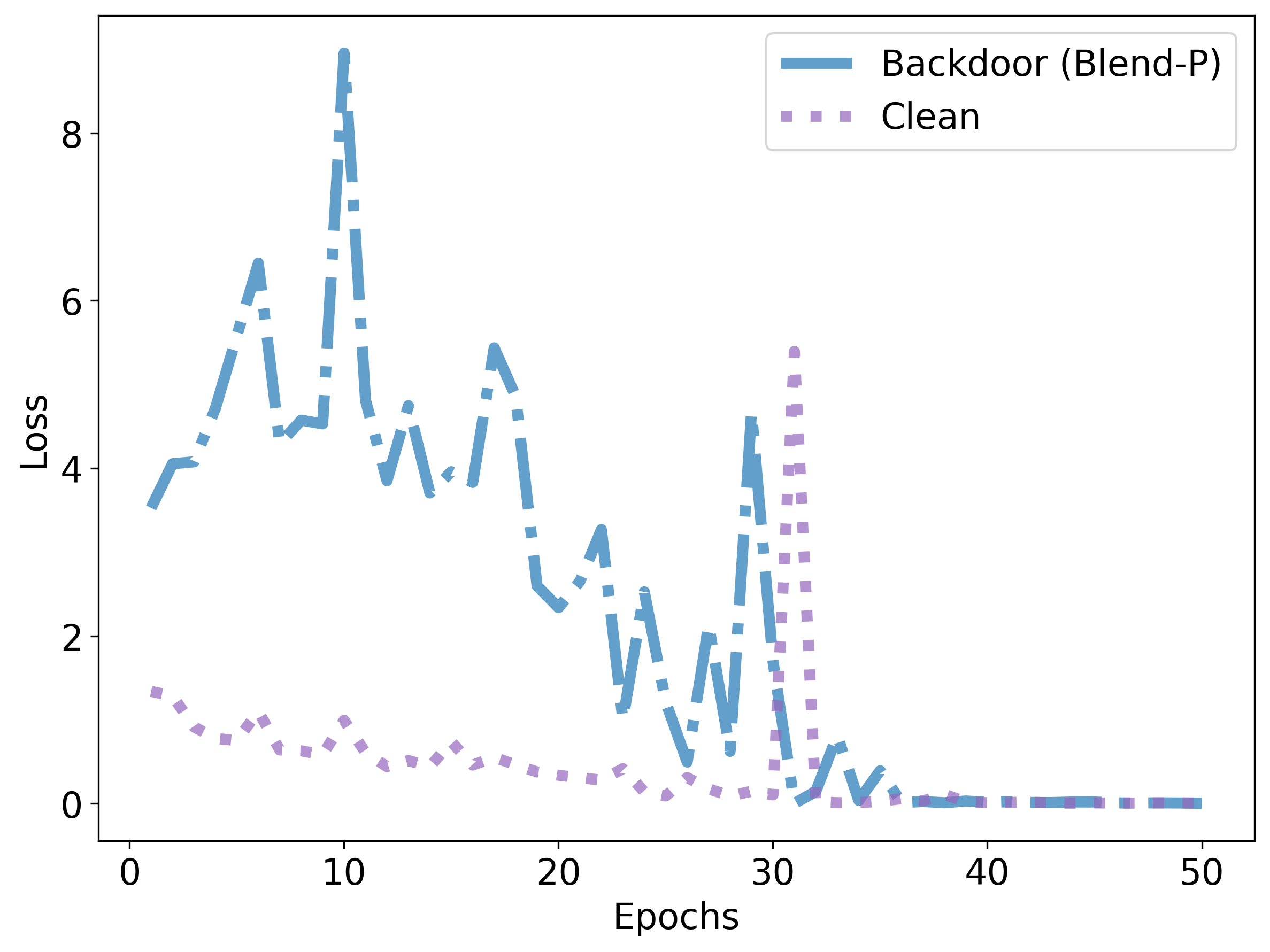}
    \end{subfigure}
    \begin{subfigure}[b]{\trajfigsize\textwidth}
         \centering
        \includegraphics[width=\textwidth]{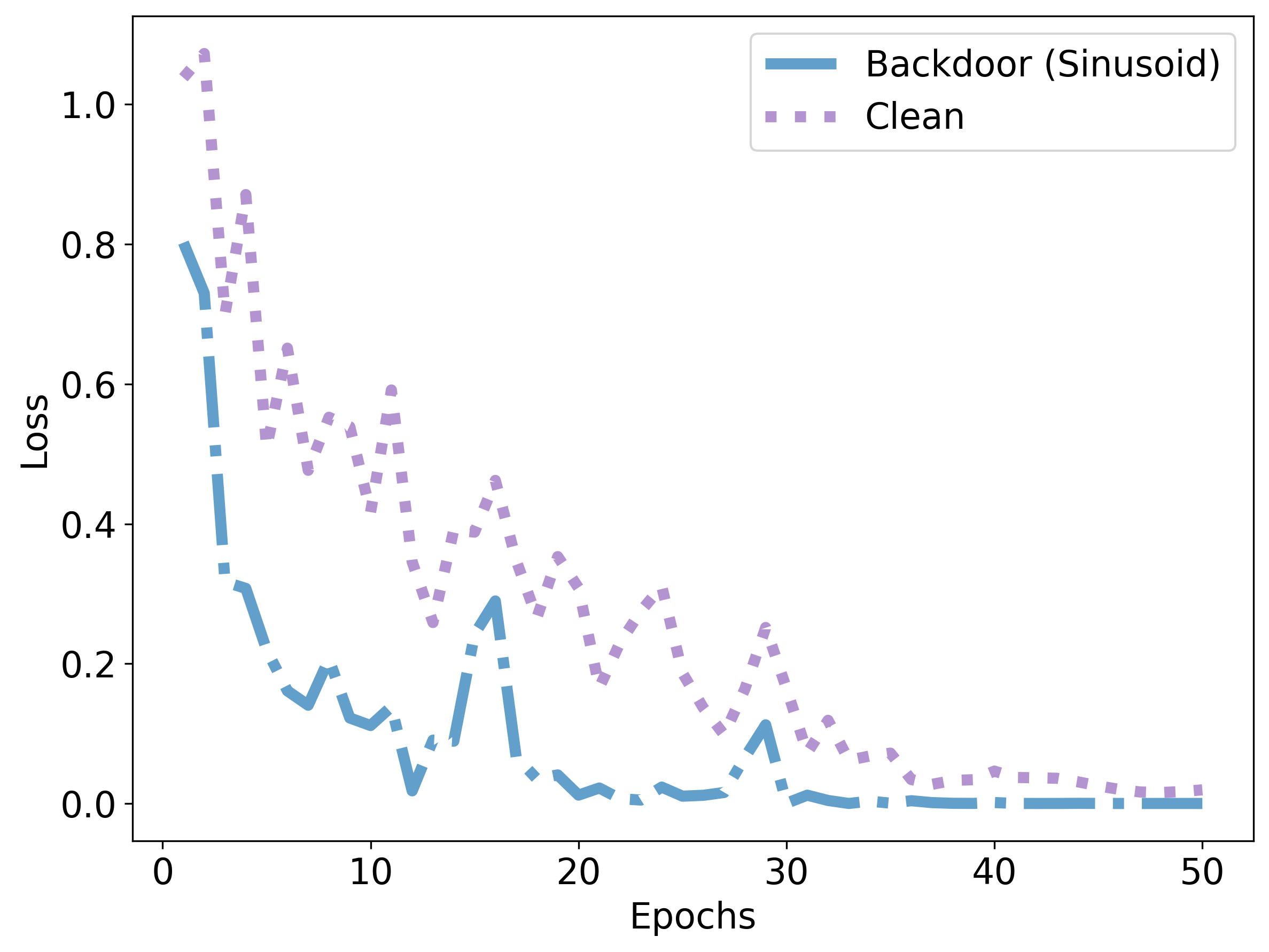}
    \end{subfigure}
    \begin{subfigure}[b]{\trajfigsize\textwidth}
         \centering
        \includegraphics[width=\textwidth]{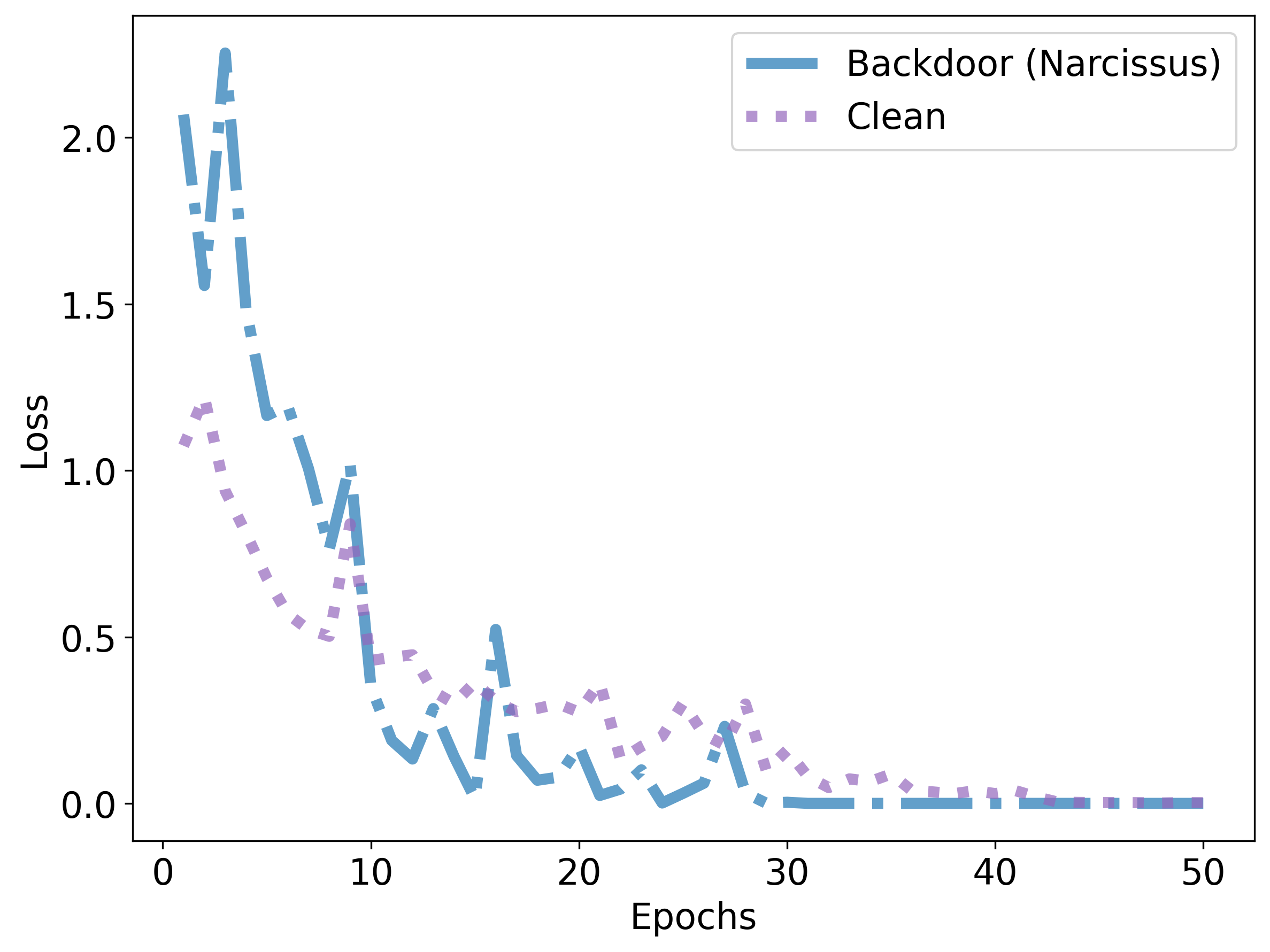}
    \end{subfigure}
    \begin{subfigure}[b]{\trajfigsize\textwidth}
         \centering
        \includegraphics[width=\textwidth]{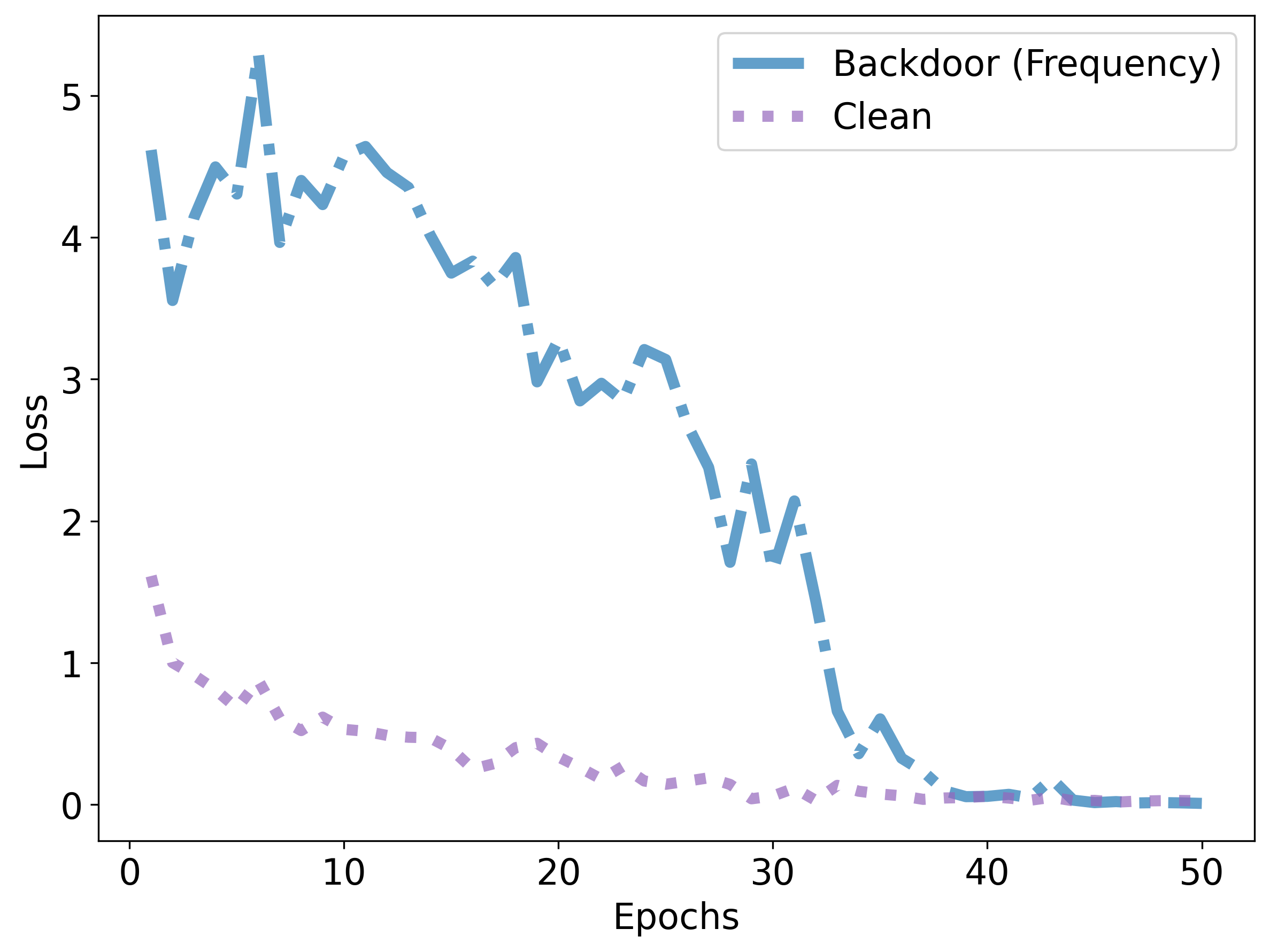}
    \end{subfigure}
    \vspace{-3mm}
    \caption{\small Average clean trajectory compared to average attack trajectories in CIFAR-10, single-attack setting, 50 epochs. \textbf{Top row, left to right:} 4-pixel patch, single-pix patch, random blended, fixed pattern blended. \textbf{Bottom row, left to right:} Sinusoid blended, narcissus, frequency attack. All backdoor attacks exhibit distinct learning dynamics from clean examples.  However, some are learned faster while others are learned slower, making the inductive bias of previous methods \citep{li2021antibackdoor, khaddaj2023rethinking, hayase2021spectre} inappropriate for defending against general poisoning attacks.}
    \label{fig:trajs}
\end{figure*}

BaDLoss relies on the observation that backdoor triggers require the model to learn features absent in a typical dataset. These features are always aberrant, as the attacker must be able to control and inject the feature into an arbitrary image in order to successfully control the backdoored model after deployment. Consequently, such examples exhibit distinct loss dynamics, illustrated in Fig.~\ref{fig:trajs}. Unlike previous defenses, BaDLoss uses entire loss trajectories to identify backdoor samples.

The BaDLoss algorithm is visualized in Fig.~\ref{fig:badloss_overview}. Given a small set of bona fide clean training examples called \textbf{probes}, the target model is trained for a small number of epochs on the target dataset $\mathcal{D}$, and the loss $\ell$ of every example is tracked at every epoch.\footnote{In the multi-attack setting, the average training loss often undergoes large spikes, visible in Figure~\ref{fig:k-trajectories} -- to mitigate this, we do not track any epoch whose average training loss is higher than twice the moving average of the last three (non-rejected) epochs.}
\begin{equation}
    \textbf{s}_{i}^{t} := \left( \ell(x_i, y_i ; \theta_{1}), \ell(x_i, y_i ; \theta_{2}), ..., \ell(x_i, y_i ; \theta_{t}) \mid (x_i, y_i) \in \mathcal{D} \right)
\end{equation}
\noindent where $\textbf{s}_i$ represents the loss trajectory for example $i^{th}$ example: ($x_i, y_i$), and $\theta_{t}$ represents the weights of the network at iteration $t$.
Treating each example's loss trajectory as a vector, we calculate a score for each example as the log of the mean $\ell_2$ distance between the target trajectory and the nearest bona fide clean examples.
\begin{equation}
    p(b \mid \textbf{s}_{i}) \propto \sum_{\textbf{g} \; \in \; \textsc{NN}(\textbf{s}_{i}, \mathcal{D}_\textbf{c}, k)} || \textbf{s}_{i}^{t} - \textbf{g} ||_2
\end{equation}
\noindent where $p(b \mid \textbf{s}_{i})$ represents the probability of the example being backdoored given the training trajectory, and $\mathcal{D}_\textbf{c}$ represents the probe dataset which is comprised of the clean bona fide examples in our case.
We discard a percentage of examples with the highest average distance from the clean trajectories. Once the backdoor examples are identified, the model is subsequently retrained on the original dataset with the backdoored examples removed.

Several hyper-parameters must be selected: $n_{clean}$, the number of bona fide clean training examples; $k$, the number of clean nearest neighbors examples to score with; $n_{epochs}$, the number of training epochs in the detection phase; and $r$, the fraction of samples to reject.

\section{Results}

\subsection{Multi-Attack Result}

\begin{figure*}[t]
    \centering
        \includegraphics[width=0.9\textwidth]{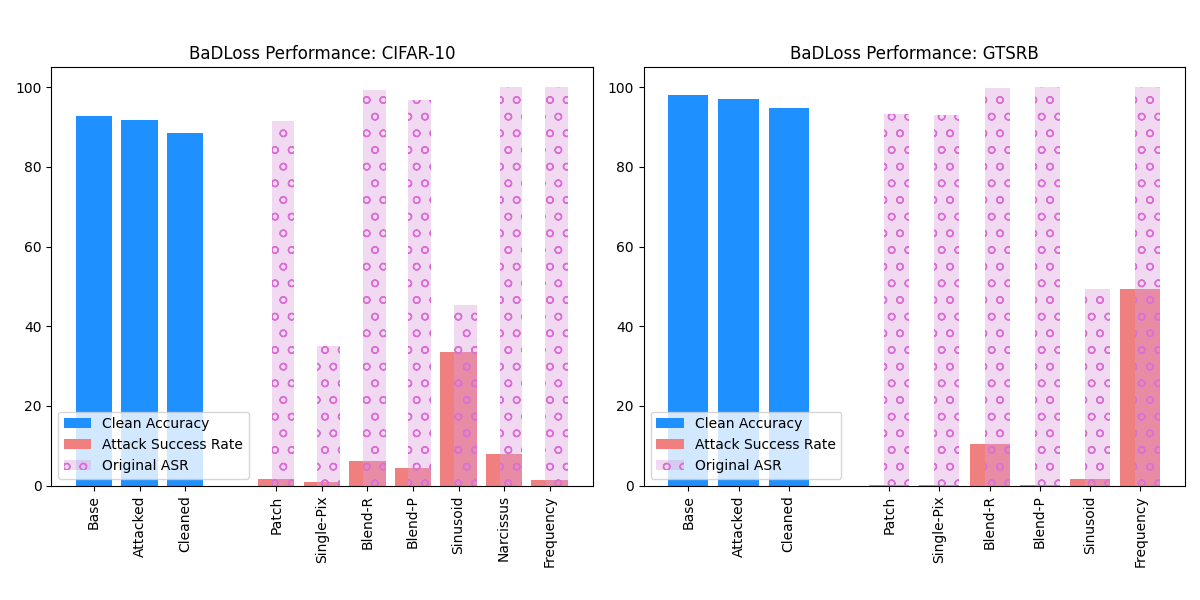}
    \caption{\small 
    \textbf{Evaluation of BaDLoss in the Multi-attack setting.}
    BaDLoss achieves very low attack success rate across all-but-one attack in both datasets, in every case substantially lowering the attack success rate.
    Contrast with Figure~\ref{fig:multiattack-defense}, which shows the failure of existing defenses.}
    \label{fig:multiattack-badloss}
\end{figure*}

Throughout training, we set $(n_{clean}, k, n_{epochs}, r) = (250, 50, 30, 0.4)$: we always track 250 bona fide clean examples and use the 50 nearest ones to calculate an example's anomaly score. We pretrain for 30 epochs, as this is usually sufficient for backdoored examples to produce different loss trajectories without becoming too computationally expensive. We always reject 40\% of training examples with the greatest distance to the clean trajectories.

The overall fraction of the dataset which is poisoned is approximately 8\% in CIFAR-10 and 10\% in GTSRB (due to needing to increase the strength of some attacks so that they are learned).

Our results are visualized in Figure~\ref{fig:multiattack-badloss} (exact numerical results available in Table~\ref{table:multi-attack} in the Appendix). BaDLoss is the first defense to credibly defend in the multi-attack setting, substantially reducing ASR on every attack considered. In both settings, BaDLoss defends against nearly every attack with only minor clean accuracy degradation.

\subsection{Single-Attack Results}

Other methods have primarily been evaluated in the single-attack setting. In order to adequately compare BaDLoss against other methods, we also evaluate its performance in the single-attack setting. Our results are visualized in Table~\ref{table:one-attack}.We demonstrate minimal sensitivity to attacks -- attaining near-zero attack sensitivity in CIFAR-10, and performing comparably well to other defense methods in GTSRB. While the high removal fraction $r=0.4$ is suitable to the multi-attack setting where the overall poisoning ratio is higher, here it degrades our clean accuracy relative to other methods. Nonetheless, these results show that our method is effective even in the single-attack setting.

\begin{table*}[t]

\begin{adjustbox}{width=0.8\linewidth,center}
\centering
\begin{tabular}{llccccccccccccccccc}
\toprule
 && \multicolumn{8}{c}{Clean Accuracy} & \phantom{abc} & \multicolumn{8}{c}{Attack Success Rate} \\
\cmidrule(lr){3-10} \cmidrule(lr){12-19}
 && \begin{sideways}Patch\end{sideways} & \begin{sideways}Single-Pix\end{sideways} & \begin{sideways}Blend-R\end{sideways} & \begin{sideways}Blend-P\end{sideways} & \begin{sideways}Sinusoid\end{sideways} & \begin{sideways}Narcissus\end{sideways} & \begin{sideways}Frequency\end{sideways} & \begin{sideways}Avg. CA\end{sideways}&
 
 & \begin{sideways}Patch\end{sideways} & \begin{sideways}Single-Pix\end{sideways} & \begin{sideways}Blend-R\end{sideways} & \begin{sideways}Blend-P\end{sideways} & \begin{sideways}Sinusoid\end{sideways} & \begin{sideways}Narcissus\end{sideways} & \begin{sideways}Frequency\end{sideways} & \begin{sideways}Avg. ASR\end{sideways}\\
\midrule
\multirow{8}{*}{\rotatebox{90}{\textbf{CIFAR-10}}}
& No Defense & 92.86 & 92.11 & 92.71 & 92.40 & 92.89 & 87.57 & 92.67 & 91.89 &  & 92.93 & 67.12 & 99.99 & 99.88 & 18.90 & 75.60 & 99.97 & 79.20 \\
& Neural Cleanse & 92.29 & 92.11 & 92.71 & 92.84 & 92.56 & 93.07 & 92.45 & \textbf{92.58} &  & 91.91 & 67.12 & 99.99 & 2.48 & 39.89 & 87.06 & \textbf{0.63} & 55.58 \\
& Activation Clustering & 90.55 & 91.96 & 92.92 & 92.76 & 92.85 & 92.07 & 92.21 & 92.19 &  & 86.89 & 54.57 & 99.91 & 99.87 & 30.86 & 91.03 & \textbf{0.63} & 66.25 \\
& Spectral Signatures & 91.64 & 90.68 & 91.50 & 91.70 & 90.68 & 90.72 & 90.98 & 91.13 &  & 88.62 & 1.32 & 1.82 & \textbf{1.87} & 23.46 & 33.89 & 0.86 & 21.69 \\
& Frequency Analysis & 31.66 & 32.52 & 36.56 & 33.89 & 21.14 & 44.79 & 32.14 & 33.24 &  & 39.78 & 12.60 & 3.83 & 5.20 & \textbf{0.50} & 2.36 & 5.02 & 9.90 \\
& Anti-Backdoor Learning & 90.33 & 90.95 & 91.83 & 91.29 & 91.33 & 90.73 & 91.49 & 91.14 &  & 92.36 & 57.91 & 99.90 & 99.87 & 39.07 & 81.59 & 98.68 & 81.34 \\
& Cognitive Distillation & 91.48 & 91.87 & 92.29 & 91.47 & 91.73 & 90.33 & 91.81 & 91.57 &  & 91.01 & 49.59 & 84.99 & 1.96 & 41.42 & 83.99 & 0.83 & 50.54 \\
& BaDLoss & 85.49 & 84.78 & 85.22 & 83.98 & 84.72 & 84.91 & 86.25 & 85.05 &  & \textbf{1.02} & \textbf{0.92} & \textbf{1.66} & 3.07 & 19.92 & \textbf{0.72} & 0.82 & \textbf{4.02} \\
\midrule
\multirow{8}{*}{\rotatebox{90}{\textbf{GTSRB}}}
& No Defense & 97.02 & 95.80 & 97.93 & 98.22 & 97.89 & - & 98.94 & 97.63 &  & 90.37 & 1.00 & 98.50 & 99.96 & 56.95 & - & 99.67 & 74.41 \\
& Neural Cleanse & 94.83 & 95.80 & 96.82 & 97.82 & 97.89 & - & 98.21 & 96.90 &  & 0.36 & 1.00 & \textbf{0.00} & \textbf{0.00} & 56.95 & - & \textbf{0.00} & \textbf{9.72} \\
& Activation Clustering & 97.16 & 95.13 & 98.17 & 97.69 & 98.16 & - & 97.51 & 97.30 &  & \textbf{0.16} & 1.03 & 97.26 & 100.00 & 59.92 & - & 98.64 & 59.50 \\
& Spectral Signatures & 96.75 & 96.06 & 97.32 & 97.69 & 97.47 & - & 97.70 & 97.17 &  & 0.55 & 0.80 & 97.04 & 100.00 & \textbf{6.84} & - & 100.00 & 50.87 \\
& Frequency Analysis & 95.27 & 95.50 & 97.95 & 98.04 & 97.93 & - & 97.89 & 97.10 &  & 1.36 & 0.74 & 98.58 & 99.82 & 66.42 & - & 100.00 & 61.15 \\
& Anti-Backdoor Learning & 93.56 & 97.77 & 95.55 & 97.28 & 96.41 & - & 97.13 & 96.28 &  & 1.97 & 2.37 & 98.62 & 99.99 & 61.11 & - & 8.62 & 45.45 \\
& Cognitive Distillation & 97.02 & 96.31 & 97.88 & 97.51 & 97.47 & - & 97.54 & 97.29 &  & 74.71 & 0.53 & 98.53 & 99.66 & 45.49 & - & 99.51 & 69.74 \\
& BaDLoss & 94.34 & 96.19 & 94.82 & 93.94 & 93.90 & - & 94.71 & 94.65 &  & 0.31 & \textbf{0.04} & 67.26 & 34.04 & 60.70 & - & 33.83 & 32.70 \\
\bottomrule
\end{tabular}
\end{adjustbox}

\vspace{5pt}
\caption{\small \textbf{Single-attack setting: Clean accuracy and attack success rate after retraining on CIFAR-10 and GTSRB.} BaDLoss is highly effective at removing detected backdoor instances, usually minimizing the backdoor's efficacy without degrading performance more than necessary.}
\label{table:one-attack}\vspace{-15pt}
\end{table*}

\section{Discussion}
\label{sec:discussion}

\subsection{Impact of the Poisoning Ratio}

As BaDLoss leverages the training dynamics of the model, poisoning ratios are highly impactful. These poisoning ratios substantially affect the loss dynamics of the corresponding examples. For instance, the random blending attack, which in CIFAR-10 is substantially harder to learn than ordinary training examples at a low poisoning ratio, instead becomes easier to learn when the poisoning ratio is increased sufficiently high. As BaDLoss detects deviations from bona fide training examples, this makes BaDLoss robust against different poisoning ratios. However, BaDLoss will struggle to detect backdoors if the poisoning ratio and attack type are chosen such that the training dynamics of the backdoored examples match the training dynamics of the bona fide training examples.

\subsection{Interactions in the Multi-attack setting}

Using multiple simultaneous attacks makes loss trajectories less stable as the model has more potential classification mechanisms to learn~\citep{lubana2023mechanistic}. As BaDLoss relies on loss trajectories, this instability impacts its performance to some extent. Despite the instabilities of individual loss trajectories, the trajectories of clean and backdoored training examples are usually distinct enough to enable identification.

Interestingly, simultaneous attacks can interact positively and improve each others' performance. Notably, in GTSRB, the single-pixel attack fails to generalize at all in the single-attack setting (and every defense trivially defends against it). However, in the multi-attack setting, the attack generalizes. We hypothesize this is due to the simultaneous patch attack in the same corner of the image. While the patch and single-pixel attacks target different classes, the patch attack teaches the network to attend to the bottom-right corner, potentially helping learn the single-pixel attack. We strongly recommend future work in model backdoors, whether attacks or defenses, to evaluate in the multi-attack setting to more fully understand how these interactions might occur in real-world environments.

\subsection{Compatibility with Other Cleaning Methods}

As we identify poisoned examples similarly to other identify-then-clean methods such as Anti-Backdoor Learning, our identification method could equivalently be used with cleaning methods other than removal-and-retrain. A full comparison of which removal methods are most effective given fixed budgets of clean and poisoned samples identified is deferred to future work.

\subsection{Limitations}
\label{sec:limitations}

We highlight some of the major limitations of our work.

\noindent\textbf{Impact of example removal.}
Our defense methodology removes examples from the training set. This is a common practice in past work~\citep{tran2018spectral}, but removal can have outsized negative impacts on the long-tail of minority classes~\citep{feldman2020memorization,liu2021jtt,sanyal2022unfair}.
Specifically, BaDLoss may more likely mark minority-class data as anomalous, as the reduced amount of training data may render trajectories unstable. Consequently, BaDLoss may exacerbate existing weaknesses on minority-class data.

\noindent\textbf{Counter-attacks.}
Generally, our assumption that attacks demonstrate anomalous loss dynamics is more robust than other models of attacks implied by other defenses (for instance, neural cleanse's model that an attack uses a low-magnitude trigger). However, an informed attacker can still exploit the design of the defense mechanism when given access to the training dataset, e.g. by carefully selecting injected features, poisoning ratios, and other hyperparameters to ensure that their attack demonstrates training dynamics that closely match the bona fide clean training examples. For instance, trained attack images (such as the sleeper agent attack~\citep{souri2022sleeper}) could add a regularization term in training to ensure that their losses mimic clean examples. 

\noindent\textbf{Self-supervised learning.}
\citet{li2023contrastivebackdoor} observe that while the training dynamics of clean and backdoored examples are distinct in supervised learning, their dynamics grow much more similar when performing self-supervised contrastive learning. BaDLoss is likely to function far more poorly in the contrastive learning, highlighting the importance of further work to develop robust defenses against multi-attack scenarios for a broader set of target domains.

\vspace{-5pt}
\section{Conclusion}
Our work makes two central contributions.
First, we introduce the problem of defending against \textbf{multiple simultaneous data poisoning attacks}. We argue this is a more realistic threat model for ML systems trained on large datasets of internet data, and show that existing defences fail in this setting.
Our second contribution is \textbf{BaDLoss}, a novel backdoor detection method which successfully defends against multiple simultaneous poisoning attacks with minimal degradation in clean accuracy.
BaDLoss works by 1) comparing the loss trajectories of individual training examples to those of a small set of known-to-be-clean examples, 2) filtering out those examples with anomalous trajectories, and 3) retraining from scratch on the filtered training set.
Our experiments demonstrate the effectiveness of BaDLoss in both single-attack and multi-attack settings.
We hope our work will inspire further research on simultaneous data poisoning attacks and believe BaDLoss provides a strong baseline for comparison.

\section{Acknowledgements}

We would like to thank Mantas Mazieka and Nicolas Papernot for their helpful guidance and feedback, as well as anonymous reviewers who provided abundant useful feedback and references.

This research was supported by the Center for AI Safety Compute Cluster. Any opinions, findings, and conclusions or recommendations expressed in this material are those of the author(s) and do not necessarily reflect the views of the sponsors.

\bibliography{refs}
\bibliographystyle{plainnat}

\newpage
\appendix

\begin{algorithm*}[t]
\DontPrintSemicolon
\SetAlgoLined
\SetKwInOut{Input}{Input}
\SetKwInOut{Output}{Output}
\SetKwInOut{Parameter}{Parameter}
\SetKwInOut{Initialize}{Initialize}
\SetKwProg{Fn}{Function}{:}{}
\SetKwProg{Class}{Class}{}{}
\SetKw{KwTo}{in}
\SetKwFor{For}{for}{do}{endfor}

\Input{Training data $X$, Labels $Y$, $n_{clean}$ bona fide clean data points $X_{c} \in X$,  Number of detection epochs $n_{epochs}$, parameter $k < n_{clean}$, rejection fraction $r$}
\Output{Clean model, Loss trajectories for all training examples, Anomaly scores}
\Parameter{Learning rate $\alpha$, Loss function $\mathcal{L}$}
\BlankLine
\Initialize{Anomaly scores $S[]$ array for per-example anomaly scores.}
\Initialize{Average losses $A[]$ array for rejecting epochs with loss spikes.}
\Initialize{Model parameters $\Theta$, Loss trajectories $T[] \leftarrow$ empty list of lists}
\For{epoch $\leftarrow 1$ \KwTo $n_{epochs}$}{
    \tcp{Train the model for one epoch on the dataset, getting the average epoch loss.}
    $\theta, l \leftarrow$ train\_model$(X, Y, \theta, \alpha, \mathcal{L})$ \;
    
    If ($l$ > 2 * average(A[-3:])): continue \;
    $A$.append(l) \;
    \tcp{Collect loss trajectory at the end of the epoch}
    T.append([])\;
    
    \For{each $(x, y)$ \KwTo $(X, Y)$}{
        predictions $\leftarrow$ model.forward(x) \tcp{no gradients are calculated}
        
        loss $\leftarrow$ $\mathcal{L}$(predictions, y)\;
        
        T$[$epoch$]$.append(loss)\;
    }
        
}

\tcp{Calculate anomaly scores}
\For{each $x$ \KwTo $X \backslash X_c$}{
    nearest\_clean\_neighbors $\leftarrow$ nearest\_neighbors($T[x], T[X_c], k$) \tcp{get n\_2 nearest clean trajectories}
    distances $\leftarrow$ distance\_metric($x$, nearest\_clean\_neighbors)\;

    $S$.append(average(distances))\;
}
\BlankLine
anomaly\_scores $\leftarrow$ S 

\tcp{Any clean training method can be used}
\tcp{Typically, removing examples below a threshold works well}
model = clean\_train($X, Y, r$, anomaly\_scores) 

\textbf{Output} model, $T$, anomaly\_scores\;

\textbf{END}\;

\label{algo:pytorch_pseudocode}
\caption{PyTorch pseudocode for BaDLoss.}
\end{algorithm*}

\section{Training Details}
\label{sec:train-details}
We use the Res-Net 50 \citep{he2015resnet} architecture throughout all our experiments. We use the AdamW optimizer with learning rate $\gamma = 1\text{e}-3$, weight decay $\lambda = 1\text{e}-4$, and $\beta_1, \beta_2 = 0.9, 0.999$. We train for 100 epochs except where otherwise specified (i.e. for BaDLoss's initial training phase of 30 epochs). In CIFAR-10, we use a batch size of 128. In GTSRB, we use a batch size of 256. In CIFAR-10, we use a crop-and-pad (4px max) and random horizontal flip augmentation except during BaDLoss pretraining. In GTSRB, we use no augmentations.

We train using PyTorch \citep{pytorch}. Our nearest neighbors classifier uses scikit-learn \citep{scikit-learn}. Plots were generated with Matplotlib \citep{matplotlib}.

\begin{figure*}[t]
    \centering
        \includegraphics[width=0.9\textwidth]{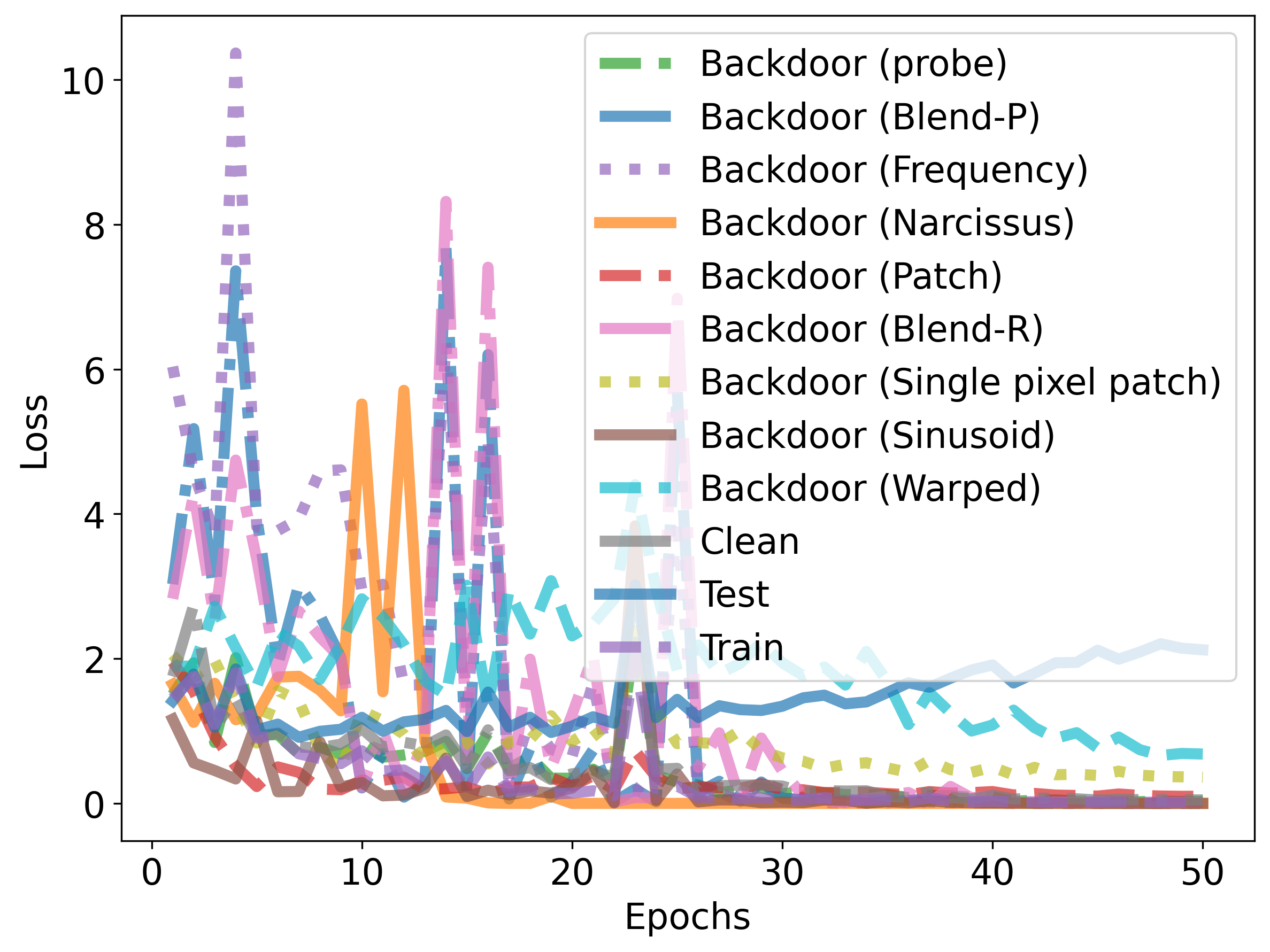}
    \caption{\textbf{Loss trajectories in the multiattack setting.}}
    \label{fig:k-trajectories}
\end{figure*}

\begin{table*}[t]

\begin{adjustbox}{width=1.0\linewidth,center}
\centering
\begin{tabular}{llcccccccccc}\toprule
 && Clean Acc. & \phantom{abc} & \multicolumn{7}{c}{Attack Success Rate} \\
\cmidrule(lr){3-3} \cmidrule(lr){5-12}
 & & & & \begin{sideways}Patch\end{sideways} & \begin{sideways}Single-Pix\end{sideways} & \begin{sideways}Blend-R\end{sideways} & \begin{sideways}Blend-P\end{sideways} & \begin{sideways}Sinusoid\end{sideways} & \begin{sideways}Narcissus\end{sideways}& \begin{sideways}Frequency\end{sideways}& \begin{sideways}Avg. ASR\end{sideways}\\
\midrule
\multirow{8}{*}{\rotatebox{90}{\textbf{CIFAR-10}}}

 & No Defense & 91.76 &  & 91.58 & 34.97 & 99.27 & 96.74 & 45.36 & 99.93 & 99.98 & 81.12 \\
 & Neural Cleanse & 91.76 &  & 91.58 & 34.97 & 99.27 & 96.74 & 45.36 & 99.93 & 99.98 & 81.12 \\
 & Activation Clustering & 91.83 &  & 92.61 & 49.24 & 99.44 & 99.21 & 44.29 & 98.04 & 99.79 & 83.23 \\
 & Spectral Signatures & 90.82 &  & 91.50 & 3.58 & 3.31 & 99.80 & 31.48 & 94.66 & 0.20 & 46.36 \\
 & Frequency Analysis & 9.84 &  & 65.34 & 13.94 & 0.00 & 66.67 & 53.69 & 0.00 & 0.00 & 28.52 \\
 & Anti-Backdoor Learning & 90.73 &  & 94.00 & 71.29 & 99.14 & 99.90 & 62.86 & 4.99 & 14.92 & 63.87 \\
 & Cognitive Distillation & 91.04 &  & 91.06 & 41.63 & 98.56 & 95.13 & 43.72 & 0.00 & 99.96 & 67.15 \\
 & BaDLoss & 88.43 &  & 1.71 & 0.92 & 6.16 & 4.29 & 33.52 & 7.81 & 1.44 & 7.98 \\

\midrule
\multirow{8}{*}{\rotatebox{90}{\textbf{GTSRB}}}

 & No Defense & 97.09 &  & 93.22 & 93.11 & 99.74 & 100.00 & 49.36 & - & 100.00 & 89.24 \\
 & Neural Cleanse & 97.09 &  & 93.22 & 93.11 & 99.74 & 100.00 & 49.36 & - & 100.00 & 89.24 \\
 & Activation Clustering & 97.82 &  & 94.96 & 94.32 & 97.05 & 100.00 & 46.69 & - & 100.00 & 88.84 \\
 & Spectral Signatures & 96.56 &  & 91.91 & 91.14 & 63.60 & 100.00 & 38.08 & - & 100.00 & 80.79 \\
 & Frequency Analysis & 97.43 &  & 94.36 & 94.17 & 0.81 & 100.00 & 54.14 & - & 100.00 & 73.91 \\
 & Anti-Backdoor Learning & 97.13 &  & 92.83 & 93.93 & 52.93 & 100.00 & 77.52 & - & 100.00 & 86.20 \\
 & Cognitive Distillation & 97.10 &  & 92.67 & 92.89 & 86.16 & 100.00 & 31.29 & - & 87.36 & 81.73 \\
 & BaDLoss & 94.75 &  & 0.13 & 0.04 & 10.32 & 0.18 & 1.75 & - & 49.32 & 10.29 \\
 
\bottomrule
\end{tabular}
\end{adjustbox}

\vspace{5pt}
\caption{\textbf{Multi-attack setting: Clean accuracy and attack success rate after retraining on CIFAR-10 and GTSRB.} This table shows that the multi-attack setting is substantially harder than the single-attack setting. BaDLoss demonstrates the best overall defense in both settings, but suffers some clean accuracy degradation in CIFAR-10.}
\label{table:multi-attack}
\end{table*}

\section{Attack and Defense Details}

\subsection{Attacks Considered}
\label{sec:attacks}

We focus on attacks that exclusively impact the training dataset. While other attacks exist, our method is not designed to defend against such threat models. Fig~\ref{fig:attacks} illustrates the attacks considered in this study. Note that attacks have very different poisoning ratios $p$: for example on CIFAR-10, ranging from 0.0005 (25 images) to 0.03 (1500 images).

\noindent\textbf{Image Patch.}
\citet{gu2017badnets} use a 4-pixel checkerboard (Patch) and a 1-pixel (Single-Pix) attack, adding a pattern and changing the corresponding image's label to the target class. We use $p=0.01$ for both attacks in CIFAR-10, and $p=0.02$ for the checkerboard patch and $p=0.04$ for the single-pixel patch in GTSRB, to make the attack more readily learnable with the larger image size.

\noindent\textbf{Blended Pattern.}
A blended pattern attack adds a full-image trigger $t$ into an image with some fraction $\alpha$, such that the attacked image $x_{\text{attacked}} = \alpha t + (1 - \alpha) x_{\text{original}}$. We use three blending attacks: \citet{chen2017backdoor}'s random noise blending (Blend-R) with $\alpha=0.075, p=0.01$,  \citet{liao2018invisible}'s dimple pattern blending (Blend-P) with $\alpha=0.025, p=0.01$, \citet{barni2019sinusoid}'s sinusoid pattern blending (Sinusoid) with $\alpha=0.075, p=0.01$.  The sinusoid attack is a clean-label baseline, so images are only selected from the target class and their labels are not changed. As GTSRB has imbalanced classes, we require that the chosen target class for the sinusoid attack has at least 1,000 images, of which the sinusoid backdoor is applied to at least 300. 

\noindent\textbf{Learned Trigger.}
\citet{zeng2022narcissus} optimize a trigger pattern to embed a backdoor with very few backdoor examples. We use $p=0.0005$ in CIFAR-10, and do not use this attack in GTSRB.

\noindent\textbf{Frequency Attack.}
\citet{wang2021frequencybackdoor} inject a trigger into an image's discrete cosine transform, acting as a high-frequency trigger pattern. We use this attack with $m=30, p=0.01$.

\subsection{Defenses Considered}
\label{sec:defenses}
We prioritize selecting defenses which are primarily filtering-based. Where possible, we adapt defenses to remove and retrain from scratch for consistence.

\noindent\textbf{Neural Cleanse.}
We use \citet{wand2019neuralcleanse}'s proposed filtering technique: finding the trigger, identifying triggered neurons, then filtering the training dataset by removing images with high trigger neuron activations, using their default threshold.

\noindent\textbf{Activation Clustering.}
We use \citet{chen2018activationclustering}'s silhouette score identification method to remove poisoned data, using their default parameters.

\noindent\textbf{Spectral Signatures.}
Spectral signatures always removes a fixed fraction (15\%) of datapoints that are deemed most anomalous~\citep{tran2018spectral}.

\noindent\textbf{Frequency Analysis.}
We construct the frequency detector~\citep{zeng2022frequency} training dataset using only the bona fide clean examples that BaDLoss has access to, for parity.

\noindent\textbf{Anti-Backdoor Learning.}
Rather than unlearn using identified examples~\citep{li2021antibackdoor}, which demonstrated substantial instability during our testing, we remove the bottom 15\% of examples after the pretraining phase (instead of the 1\% usually removed and used for unlearning) and retrain from scratch.

\noindent\textbf{Cognitive Distillation.}
Similarly to Anti-Backdoor Learning, instead of doing an unlearning phase after finding the lowest magnitude masks~\citep{huang2023cognitivedistillation}, we remove the fixed 15\% of examples with the lowest magnitudes and retrain from scratch.

\begin{figure*}[p]
    \centering
        \includegraphics[width=0.9\textwidth]{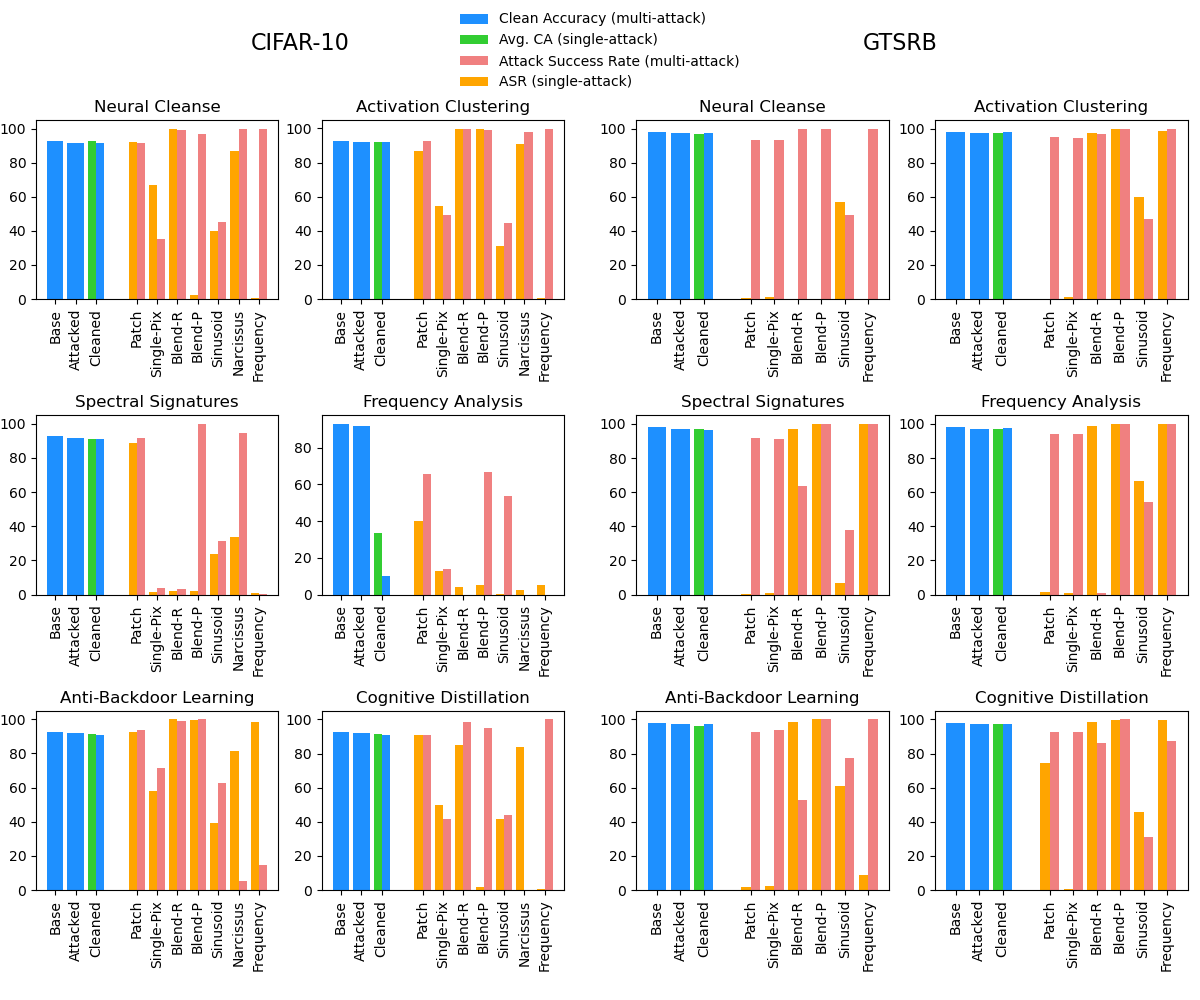}
        \includegraphics[width=0.9\textwidth]{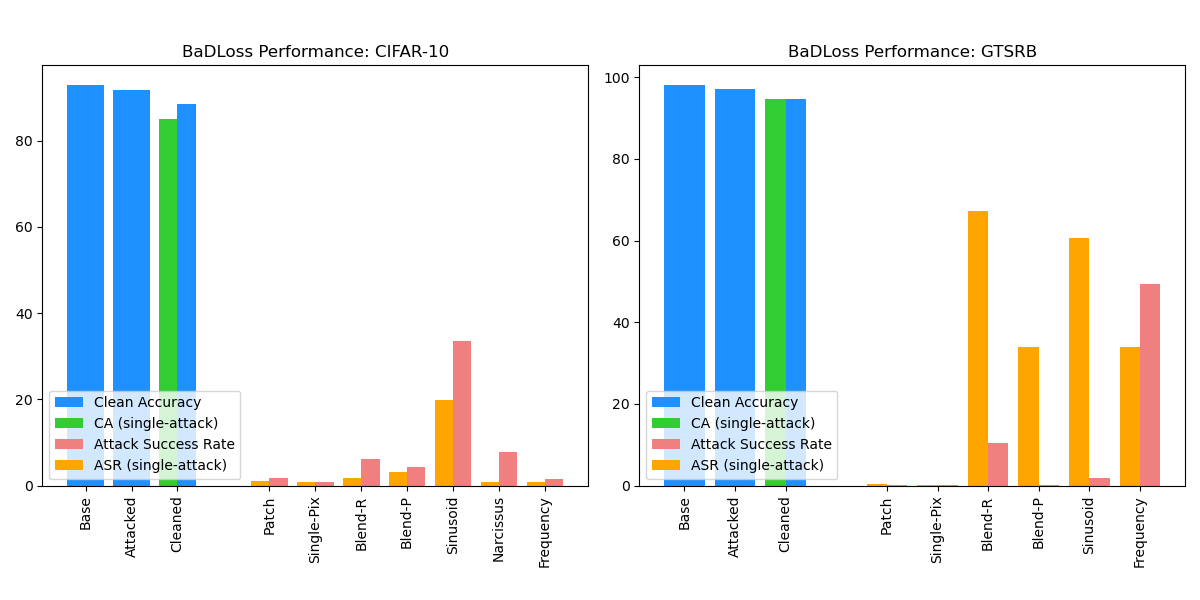}
    \caption{This figure demonstrates the difference in performance for different defenses between the single and multi-attack setting. \textbf{Top:} Existing defense performance is erratic between the single and multi-attack settings, but areas with low orange bars (attack is defended in isolation) but high red bars (attack succeeds when executed with other simultaneous attacks) are visible. \textbf{Bottom:} BaDLoss performance is highly stable regardless of single or multi-attack, and in GTSRB even improves on average in the multi-attack setting.}
    \label{fig:single-vs-multi}
\end{figure*}

\newpage
\if@preprint
\section*{NeurIPS Paper Checklist}

\begin{enumerate}

\item {\bf Claims}
    \item[] Question: Do the main claims made in the abstract and introduction accurately reflect the paper's contributions and scope?
    \item[] Answer: \answerYes{} %
    \item[] Justification: We demonstrate empirically that defenses fail in the multi-attack setting, and that our proposed method functions well.

\item {\bf Limitations}
    \item[] Question: Does the paper discuss the limitations of the work performed by the authors?
    \item[] Answer: \answerYes{}
    \item[] Justification: We have a discussion of limitations in Subsection~\ref{sec:limitations}. We do not claim that this discussion is exhaustive.

\item {\bf Theory Assumptions and Proofs}
    \item[] Question: For each theoretical result, does the paper provide the full set of assumptions and a complete (and correct) proof?
    \item[] Answer: \answerNA{}
    \item[] Justification: No theoretical results are provided.

    \item {\bf Experimental Result Reproducibility}
    \item[] Question: Does the paper fully disclose all the information needed to reproduce the main experimental results of the paper to the extent that it affects the main claims and/or conclusions of the paper (regardless of whether the code and data are provided or not)?
    \item[] Answer: \answerYes{}
    \item[] Justification: Most key information is provided. However, due to the higher variances induced by interacting attacks in the multi-attack setting, an exact reproduction would likely require code, whose provision is deferred till after the double-blind review.

\item {\bf Open access to data and code}
    \item[] Question: Does the paper provide open access to the data and code, with sufficient instructions to faithfully reproduce the main experimental results, as described in supplemental material?
    \item[] Answer: \answerNA{}
    \item[] Justification: All data used in this paper is freely available. Code is not provided, but will be provided following double-blind review.

\item {\bf Experimental Setting/Details}
    \item[] Question: Does the paper specify all the training and test details (e.g., data splits, hyperparameters, how they were chosen, type of optimizer, etc.) necessary to understand the results?
    \item[] Answer: \answerYes{}
    \item[] Justification: This is provided in Section~\ref{sec:train-details}.

\item {\bf Experiment Statistical Significance}
    \item[] Question: Does the paper report error bars suitably and correctly defined or other appropriate information about the statistical significance of the experiments?
    \item[] Answer: \answerNo{}
    \item[] Justification: No error bars are reported -- due to the breadth of experiments required, $n=1$.

\item {\bf Experiments Compute Resources}
    \item[] Question: For each experiment, does the paper provide sufficient information on the computer resources (type of compute workers, memory, time of execution) needed to reproduce the experiments?
    \item[] Answer: \answerNo
    \item[] Justification: Any compute resources can be used to achieve the training requirements set out in  Section~\ref{sec:train-details}. We have not reported compute used in experiments or in final experiments, but can do so if required.
    
\item {\bf Code Of Ethics}
    \item[] Question: Does the research conducted in the paper conform, in every respect, with the NeurIPS Code of Ethics \url{https://neurips.cc/public/EthicsGuidelines}?
    \item[] Answer: \answerYes{} %
    \item[] Justification: We satisfy the guidelines in the linked URL. Where our paper has potential negative consequences, we discuss them in our limitations (Subsection~\ref{sec:limitations}).

\item {\bf Broader Impacts}
    \item[] Question: Does the paper discuss both potential positive societal impacts and negative societal impacts of the work performed?
    \item[] Answer: \answerYes %
    \item[] Justification: The potential positive impacts are taken to be self-evident, as improving backdoor defense to function in a more realistic setting is obviously useful and societally positive. As stated above, the negative impacts are discussed in Subsection~\ref{sec:limitations}.
    
\item {\bf Safeguards}
    \item[] Question: Does the paper describe safeguards that have been put in place for responsible release of data or models that have a high risk for misuse (e.g., pretrained language models, image generators, or scraped datasets)?
    \item[] Answer: \answerNA{} %
    \item[] Justification: Nothing in this paper has a substantial risk of misuse.

\item {\bf Licenses for existing assets}
    \item[] Question: Are the creators or original owners of assets (e.g., code, data, models), used in the paper, properly credited and are the license and terms of use explicitly mentioned and properly respected?
    \item[] Answer: \answerYes{} %
    \item[] Justification: Data, models, and code used in the paper are credited to the best of the authors' knowledge.

\item {\bf New Assets}
    \item[] Question: Are new assets introduced in the paper well documented and is the documentation provided alongside the assets?
    \item[] Answer: \answerNA{} %
    \item[] Justification: The only new asset provided is our original code, which has not yet been released.

\item {\bf Crowdsourcing and Research with Human Subjects}
    \item[] Question: For crowdsourcing experiments and research with human subjects, does the paper include the full text of instructions given to participants and screenshots, if applicable, as well as details about compensation (if any)? 
    \item[] Answer: \answerNA{} %
    \item[] Justification: No human subjects were used.

\item {\bf Institutional Review Board (IRB) Approvals or Equivalent for Research with Human Subjects}
    \item[] Question: Does the paper describe potential risks incurred by study participants, whether such risks were disclosed to the subjects, and whether Institutional Review Board (IRB) approvals (or an equivalent approval/review based on the requirements of your country or institution) were obtained?
    \item[] Answer: \answerNA{} %
    \item[] Justification: No human subjects were used, so no IRB approval was required.

\end{enumerate}
\fi

\end{document}